\documentclass[letterpaper]{article} % DO NOT CHANGE THIS
\usepackage{aaai2026}  % DO NOT CHANGE THIS
\usepackage{times}  % DO NOT CHANGE THIS
\usepackage{helvet}  % DO NOT CHANGE THIS
\usepackage{courier}  % DO NOT CHANGE THIS
\usepackage[hyphens]{url}  % DO NOT CHANGE THIS
\usepackage{graphicx} % DO NOT CHANGE THIS
\usepackage{natbib}  % DO NOT CHANGE THIS AND DO NOT ADD ANY OPTIONS TO IT
\usepackage{caption} % DO NOT CHANGE THIS AND DO NOT ADD ANY OPTIONS TO IT
\usepackage{inconsolata}
\usepackage{latexsym}
\usepackage{mathtools}
\usepackage{booktabs}
\usepackage{amssymb}
\usepackage{multirow}
\usepackage{algorithm}
\usepackage{algorithmic}
\usepackage{amsmath}
\usepackage[utf8]{inputenc}
\usepackage[most]{tcolorbox}
\usepackage{fancyhdr}
\usepackage{newfloat}
\usepackage{listings}
\DeclareCaptionStyle{ruled}{labelfont=normalfont,labelsep=colon,strut=off} % DO NOT CHANGE THIS
\floatstyle{ruled}
\newfloat{listing}{tb}{lst}{}
\floatname{listing}{Listing}
\title{ALTER: Asymmetric LoRA for Token-Entropy-Guided Unlearning of LLMs}
\author {
    Xunlei Chen\textsuperscript{\rm 1, \thanks{These authors contributed equally to this work.}},
    Jinyu Guo\textsuperscript{\rm 1, *, \dag},
    Yuang Li\textsuperscript{\rm 1},
    Zhaokun Wang\textsuperscript{\rm 1, \dag},
    Yi Gong\textsuperscript{\rm 1},
    Jie Zou\textsuperscript{\rm 2},
    Jiwei Wei\textsuperscript{\rm 2},
    Wenhong Tian\textsuperscript{\rm 1, }\thanks{Corresponding author}
}

\affiliations {
    \textsuperscript{\rm 1}School of Information and Software Engineering, University of Electronic Science and Technology of China\\
    \textsuperscript{\rm 2}School of Computer Science and Engineering, University of Electronic Science and Technology of China\\
    wzk@std.uestc.edu.cn, \{guojinyu, tian\_wenhong\}@uestc.edu.cn
}

\usepackage{bibentry}
\begin{document}

\maketitle

\begin{abstract}
Large language models (LLMs) have advanced to encompass extensive knowledge across diverse domains. Yet controlling what LLMs should not know is important for ensuring alignment and thus safe use. However, effective unlearning in LLMs is difficult due to the fuzzy boundary between knowledge retention and forgetting. This challenge is exacerbated by entangled parameter spaces from continuous multi-domain training, often resulting in collateral damage, especially under aggressive unlearning strategies. Furthermore, the computational overhead required to optimize State-of-the-Art (SOTA) models with billions of parameters poses an additional barrier. In this work, we present \textbf{ALTER}, a lightweight unlearning framework for LLMs to address both the challenges of knowledge entanglement and unlearning efficiency. ALTER operates through two phases: (I) high entropy tokens are captured and learned via the shared A matrix in LoRA, followed by (II) an asymmetric LoRA architecture that achieves a specified forgetting objective by parameter isolation and unlearning tokens within the target subdomains. Serving as a new research direction for achieving unlearning via token-level isolation in the asymmetric framework. ALTER achieves SOTA performance on TOFU, WMDP, and MUSE benchmarks with over 95\% forget quality and shows minimal side effects through preserving foundational tokens. By decoupling unlearning from LLMs' billion-scale parameters, this framework delivers excellent efficiency while preserving over 90\% of model utility, exceeding baseline preservation rates of 47.8-83.6\%.
\end{abstract}

\begin{links}
\link{Code}
{https://github.com/MastrOrigami/ALTER.git}
\end{links}

\section{Introduction}
LLMs have demonstrated remarkable capabilities in downstream task processing \cite{Lee2020BioBERT, wangnoise} and content generation through unprecedented model scale expansion \cite{Achiam2023Gpt} and diversification of pretraining data growth \cite{zhao2023Survey}. However, there are significant challenges in controlling the generation of sensitive content \cite{Carlini2022Quantifying}, private information, or illegal content \cite{Shi2024Safety}. As the legal provisions of the \textit{General Data Protection Regulation} (GDPR) and the \textit{``right to be forgotten”} gain increasing attention. \cite{Grynbaum2023Times}, unlearning for LLMs has emerged as a rapidly growing research area \cite{cha2024learning}, aiming to selectively eliminate the influence of specific knowledge from deployed models \cite{yao2024unlearning, si2023knowledge}.
% Rosen2011right,

\begin{figure}[t]
\centering
\includegraphics[scale=0.38]{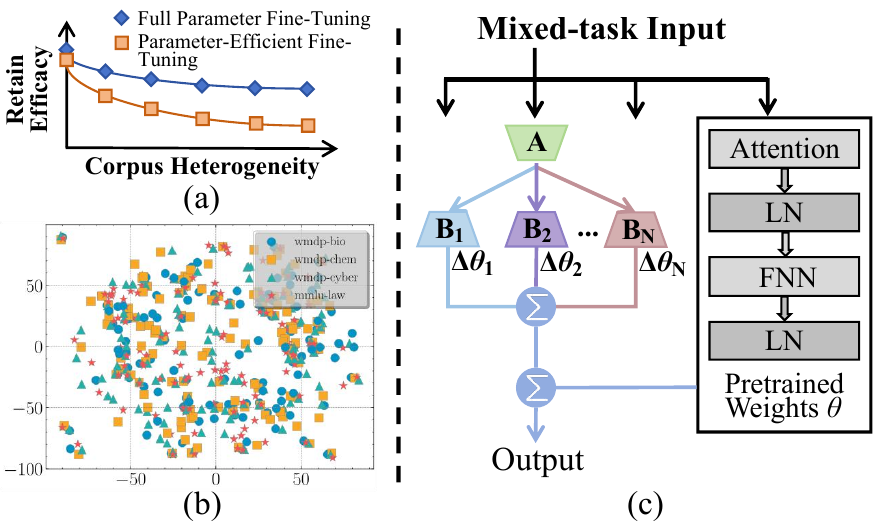}
\caption{(a) The impact of corpus heterogeneity on the performance of FT/PEFT. (b) The chaos in the LoRA parameter space caused by corpus heterogeneity in the WMDP dataset.} 
\label{intro}
\end{figure}

Current unlearning methods comprise: (1) Prompt-based \cite{liu2024large} and auxiliary model \cite{ilharco2023editing, ji2024reversing} techniques that induce forgetting without parameter modification but exhibit limited generalization, robustness; (2) Fine-tuning (FT) utilizing losses (e.g., gradient ascent on forgetting sets \cite{liu2024revisiting}) that face scalability challenges for billion-parameter models and knowledge consistency issues; (3) Model editing employing local modifications like Low-Rank Adaptation (LoRA) \cite{hu2022lora} or unlearn layers \cite{yu2025unierase}. As a leading parameter-efficient fine-tuning (PEFT) technique, LoRA optimizes merely 0.1\%-10\% parameters, accelerates training 5-20x \cite{he2021towards}, preserves knowledge integrity, and inherently enables regulatory ``Proof Unlearning'' through its modular architecture.

However, the target domains of current unlearning scenarios (Fig.\ref{intro}a) exhibit complex subdomains and task diversity. Despite forgetting strategies, residual information from heterogeneous data \cite{Carlini2021Extracting, Carlini2022Quantifying} persists via shared parameters \cite{Wang2024Unlearning}, inducing parameter coupling in both PEFT and FT (Fig.\ref{intro}b). This entanglement reduces unlearning efficacy \cite{liu2025rethinking} and risks ``over-forgetting'' \cite{Yao2024survey, Shi2024Safety}. Therefore, a natural but non-simple question arises: How can PEFT achieve efficient unlearning while preserving overall performance in multi-domain coupled parameter spaces?

Building on prior discussion, we note that asymmetric LoRA architectures offer unique advantages: prior studies \cite{tian2024hydralora} demonstrate that the shared matrix A typically captures universal knowledge while individual B matrices adapt to discrepancy knowledge (Fig.\ref{Attention}A). 

Inspired by this foundation, we propose the Asymmetric LoRA for Token-Entropy-Guided Unlearning (ALTER) framework, which achieves parameter isolation across forgetting subtasks and decoupling between forgetting and retention tasks. This design eliminates catastrophic forgetting caused by parameter entanglement in traditional methods. To precisely remove token-level sensitive knowledge, we introduce Token-Entropy-Guided: via token-wise entropy modeling, foundational tokens (high entropy) are preserved in shared matrix A, while task-specific tokens are stored in distinct B matrices. This solves public knowledge miss (grammatical tokens) from sentence-level forgetting, establishes a dynamic forgetting boundary preserving knowledge topology integrity, and enables interpretability through information entropy modeling.

ALTER achieves SOTA performance on TOFU, WMDP, and MUSE benchmarks with over 95\% forget quality. By decoupling unlearning from the billion-scale parameters of LLMs, this framework is highly efficient while preserving over 90\% of the model's utility (compared to 47.8-83.6\% for baselines), and demonstrates minimal side effects by preserving foundational tokens.

In summary, our contributions can be listed as follows:

\begin{itemize}
    \item We propose ALTER, an unlearning framework that avoids fine-tuning LLMs' weights via LoRA, achieving parameter isolation across forgetting subtasks and decoupling between forgetting and retention tasks.
    \item We introduce a token-entropy-based localization method that enables precise, selective forgetting and alleviates over-forgetting.
    \item We conduct extensive experiments across three knowledge unlearning tasks to validate forgetting quality, model utility, and fluency.
\end{itemize}

\section{Background and Motivation}
\paragraph{Mainstream Unlearning Strategies}
We model LLMs as probabilistic models $\pi_{\theta}$. Under single LoRA, the forward computation is $\mathbf{W} = \mathbf{W}_0 + \mathbf{BA}$. The forgetting set $\mathcal{D}_f$ and retention set $\mathcal{D}_r$ are comprised of question-answer pairs $(q, a)$. Mainstream parameter-directional forgetting strategies minimize $\pi_\theta$'s joint loss:
\begin{equation}
\begin{split}
\arg\min_{\pi_{\theta}}  = \beta \underbrace{\mathbb{E}_{(q,a)\sim\mathcal{D}_{f}}\left[l_{f}(q\mid a;\pi_{\theta})\right]}_{\text{forgetting term}} \\
+ \gamma \underbrace{\mathbb{E}_{(q,a)\sim\mathcal{D}_{r}}\left[l_{r}(q\mid a;\pi_{\theta})\right]}_{\text{retaining term}}.
\end{split}
\end{equation}

The loss $l_f$ prevents the model from answering questions in the forgetting set $\mathcal{D}_f$, while $l_r$ preserves base capabilities and responses to $\mathcal{D}_r$. Weighting coefficients $\beta$ and $\gamma$ balance these losses. Specific details of $l_f$ and $l_r$ are in Appendix D.

\paragraph{Asymmetric LoRA (AsymLoRA).}
The one-to-many architecture extends the single LoRA, with its formula defined as:
% with its weight update formula defined as:
\begin{equation}
\begin{split}
\mathbf{W} = \mathbf{W}_0 + \Delta \mathbf{W} = \mathbf{W}_0 + \sum_{i=1}^{N} \omega_i \cdot \mathbf{B}_i \mathbf{A}.
\end{split}
\label{EqAsymLoRA}
\end{equation}

$\mathbf{W}_0$ denotes the frozen pretrained weight matrix, while $\Delta \mathbf{W}$ represents AsymLoRA's incremental update matrix. Here, $\mathbf{A}\in\mathbb{R}^{r\times k}$ and ${\mathbf{B}_i\in\mathbb{R}^{d\times r}}(i=1,2,\dots,N)$ with weighting coefficients $\omega_i$ scaling each $\mathbf{B}_i$'s contribution, where $N$ is the number of $\mathbf{B}$ matrices and $r$ is the rank of the low-rank decomposition.

\paragraph{Token Entropy}
Shannon entropy \cite{shannon1948mathematical} quantifies predictive information in models. For LLMs, the entropy of an input token $ x_t$ is defined as:
\begin{equation}
\begin{split}
H(x_t) = -\sum_{i=1}^{V} p_{t,i} \log p_{t,i}.
\end{split}
\end{equation}

$p_{t,i} \in \mathbb{R}^V$ denotes the model's token probability at position $t$ for vocabulary index $i$. ``Token entropy'' describes the uncertainty in the probability distribution over token generation at position $t$ in LLMs, which is determined by the model's output logits at that position.

\begin{figure*}[ht]
\centering
\includegraphics[scale=0.52]{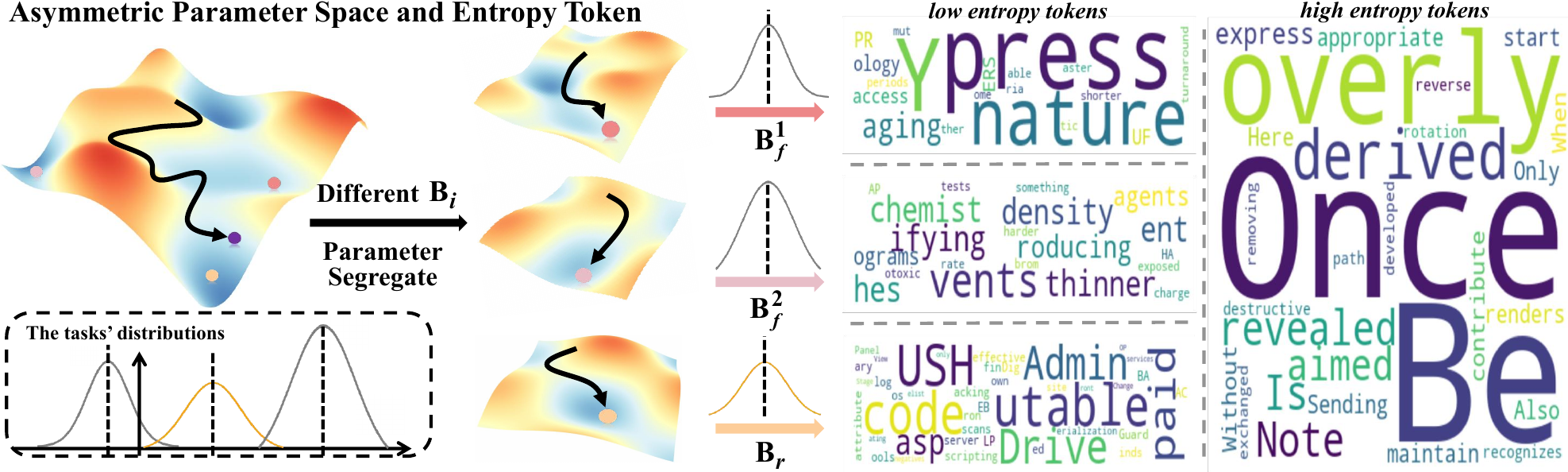}
\caption{Conceptual illustration of our unlearning framework. After achieving explicit parameter isolation with the AsymLoRA structure, word clouds from the WMDP dataset show that task-specific forgetting experts $\mathbf{B}_i$ and the retention expert $\mathbf{B}_r$ process low entropy tokens (left), whereas the shared matrix $\mathbf{A}$ processes high entropy tokens (right).} 
\label{Entropy-Token}
\end{figure*}

\paragraph{\textit{Observation I: Asymmetric LoRA efficiently achieves parameter isolation for unlearning.}}
AsymLoRA enables efficient unlearning by establishing a natural boundary between shared and task-specific knowledge (Fig.\ref{Entropy-Token}). The subset requiring removal from $\mathcal{D}_f$ is denoted as the specific forgetting sub-domain, with its optimization objective defined as:
\begin{equation}
\begin{split}
    \min_{\omega_f^d, \omega_r}\, 
    \beta \mathbb{E}_{(q,a) 
    \sim \mathcal{D}_f^d}
    \underbrace{
    \left[
    l_f\left(q \mid a; \left(\pi_{\theta} +
        \omega_f^d \cdot \mathbf{B}_d \mathbf{A}\right)\right)
    \right]
    }_{\mathclap{\text{forgetting term}}}
\\ +   
    \gamma \, \mathbb{E}_{(q,a) \sim \mathcal{D}_r}
    \underbrace{
    \left[
        l_r\left(q \mid a; \left(\pi_{\theta} + \mathbf{B}_r \mathbf{A}\right)\right)
    \right]
}_{\mathclap{\text{retaining term}}}.
\end{split}
\label{Obs1_formula}
\end{equation}

Here, $\omega_f^d$ modulates the contribution weights for head $\mathbf{B}_d$. $\mathbf{B}_r$ serves as the globally shared expert for retained knowledge. The inherent isolation mechanism transforms the complex problem of forgetting heterogeneous data into localized optimization tasks for specific data subsets. Further interpretation of parameter space is provided in Appendix F.1.

Although AsymLoRA achieves subtask-level parameter isolation \textbf{\textit{(Observation I)}}, its instance-level granularity treats QA pairs atomically, ignoring internal token heterogeneity. Uniform unlearning damages language structures (e.g., transition words like ``however''). True selective forgetting thus requires token-level intervention to precisely remove target knowledge while preserving structural tokens. This idea of precise forgetting based on token-level knowledge localization leads to the following discovery.

\paragraph{\textit{Observation II: Token-level entropy demonstrates a robust bimodal distribution aligned with linguistic functions, maintaining stability during model adaptation.}} 
The word clouds in Fig.\ref{Entropy-Token} reveal functional token separation by entropy: high entropy tokens are mostly structural elements (e.g., ``however''/``therefore''), while low entropy tokens contain knowledge-intensive content (e.g., ``entities''). This pattern (Appendix F.2) remains stable during PEFT: more than 87\% high entropy tokens retain uncertainty and more than 92\% low entropy tokens maintain determinism when fine-tuning Llama3-8B on WMDP. Entropy conservation thus enables token-level knowledge management: low entropy tokens permit precise factual removal while high entropy tokens preserve structural integrity. More token analysis is provided in Appendix F.

\begin{figure*}[ht]
\centering
\includegraphics[scale=0.55]{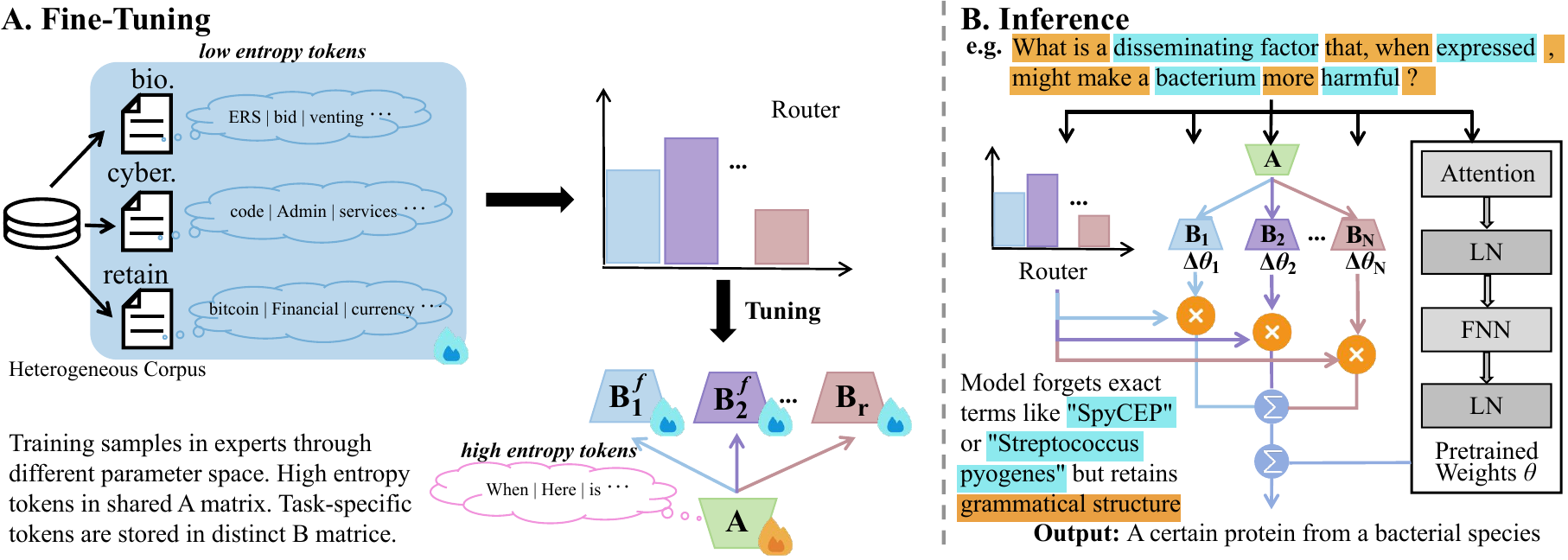}
\caption{Architecture and workflow of our unlearning framework.
During fine-tuning, ALTER first automatically identifies and initializes $N$ intrinsic components (without requiring domain-specific knowledge). Then, guided by entropy, the architecture uses a trainable MoE router that treats each intrinsic component as an expert, automatically assigning training samples to the corresponding component. High entropy tokens (red fire), inherently adapted to the shared $\mathbf{A}$ matrix, are processed jointly, while low entropy tokens (blue fire) are directed to specialized $\mathbf{B}$ experts for fine-tuning. During inference, ALTER dynamically combines multiple $\mathbf{B}$ matrices using the trained router for flexible and adaptive unlearning.} 
\label{Attention}
\end{figure*}

\paragraph{\textit{Motivation: Knowledge forgetting must simultaneously accommodate structural modularity and token-level specificity, integrating architecture isolation with entropy-driven token partitioning via Tsallis entropy.}} 
Building upon our observations, the asymmetric $\mathbf{A}$-$\mathbf{B}$ decomposition establishes vertical isolation: the shared matrix $\mathbf{A}$ inherently adapts to high entropy tokens (H$>$2.0), which are connectives or logical words that represent language structure invariants; the individual matrices $\mathbf{B}_i$ carry the domain expert knowledge contained in the low entropy tokens (Appendix F.2), where $\mathbf{B}_f^1,\dots,\mathbf{B}_f^d$ handle knowledge to be forgotten in specific sub-domains, and $\mathbf{B}_r$ strengthens knowledge in the retention set. The entropy hierarchy creates horizontal distinction (training pipeline in Fig.\ref{Attention}): high entropy tokens aggregate in $\mathbf{A}$, low entropy tokens isolate in $\mathbf{B}_i$ experts. However, Shannon entropy's independence assumption contradicts LLMs' non-extensive nature \cite{tsallis1988possible}, causing semantic coupling in $\mathbf{B}_i$ and nonlinear $\mathbf{A}$ interactions that break $\beta$-$\gamma$ trade-off in Eq.\ref{Obs1_formula}. We therefore introduce Tsallis entropy for hierarchical modeling:
\begin{equation}
\begin{split}
 S_q(x_t) = \frac{1}{q-1} \left( 1 - \sum_{i=1}^{V} p_{t,i}^q \right) \quad (q>0).
\end{split}
\end{equation}

Here, $x_t$ is the input token, typically a vector. Deformation parameter $q$ models knowledge association strength for dual control: (1) $q<1$ enhances structural invariance in $\mathbf{A}$ for high entropy tokens. (2) For $\mathbf{B}_i$ experts, $q>1$ breaks low entropy tokens' cross-domain associations, enabling forgetting in targeted domains while preventing collateral damage. Non-extensive Tsallis entropy establishes a knowledge-specific unlearning manifold between architecture isolation (physical separation of $\mathbf{B}_i$) and entropy partitioning (separation functional high $S_q$ tokens), resolving the core conflict  through consolidation of shared knowledge in the $\mathbf{A}$ matrix and targeted perturbation within $\mathbf{B}_i$ experts. Theoretical analysis is provided in Appendix A.

\section{ALTER: Asymmetric LoRA for Token-Entropy-Guided Unlearning}
\subsection{Token-Entropy-Guided Architecture}

Based on the above discussion, we propose the Token-Entropy-Guided Architecture. Building upon the isolation framework in Eq.\ref{EqAsymLoRA} above, we extend it to include a retention set:
\begin{equation}
\begin{split}
\mathbf{W} &= \mathbf{W}_0 + \Delta \mathbf{W} \\
&= \mathbf{W}_0 + \left(\mathbf{B}_r +\sum_{d=1}^{N} \omega_f^d \cdot \mathbf{B}_f^d \right)\mathbf{A}.
\end{split}
\end{equation}

Shared matrix $\mathbf{A}$ captures task-invariant knowledge, and is initialized via Kaiming as $\mathcal{N}(0, \sigma^2)$ to align high entropy tokens with pretrained weights. Experts $\mathbf{B}_f^d$ and $\mathbf{B}_r$ minimize KL divergence by mapping task-specific clustering centers to low entropy feature spaces, introducing initial heterogeneity in the parameter distribution entropy $S(\mathbf{B})$.
Specifically, each forgetting expert $\mathbf{B}_f^d$ corresponds to subdomain $d$, and is initialized from $\mathcal{D}_f^d$'s clustering center ($N$ total domains). The retention expert $\mathbf{B}_r$ preserves knowledge, and is initialized via $\mathcal{D}_r$'s feature distribution.

\subsection{Training Stage}
\paragraph{Forward Propagation}
We introduce an entropy-based adaptive gating mechanism. For input $x_t$, the function is defined as:
\begin{equation}
g_d(x_t) = \text{softmax}({W_g^T \cdot S_q(x_t)}/{\tau}).
\end{equation}

Here, $W_g^T$ denotes the transpose of a learnable weight matrix $W_g$ and $S_q(\cdot)$ represents the Tsallis entropy. The routing temperature $\tau$ is dynamically adjusted: for high entropy tokens ($S_q(x_t) > 1.2$), we set $\tau = 0.8$ to activate multiple experts and enhance structural robustness; for low entropy tokens ($S_q(x_t) \leq 1.2$), $\tau = 0.01$ is used to enforce single-expert precision routing.

\paragraph{Loss Calculation and Backpropagation}
Inspired by the classic hinge loss \cite{cortes1995support}, detailed in Appendix B, we design ALTER's hierarchical loss, extending the Eq.\ref{Obs1_formula} into a three-level cascaded optimization:
\begin{equation}
\begin{split}
\min_{\omega_f^d, \omega_r} & \quad \beta \sum_{d=1}^N \mathbb{E}_{(q,a)\sim\mathcal{D}_f^d}  \underbrace{\left[\mathcal{L}_{\text{IHL}}\left(q \mid a; \left(\pi_{\theta} + \omega_f^d \mathbf{B}_f^d\mathbf{A}\right)\right)\right]}_{\mathclap{\text{entropy-weighted forgetting}}}  \\ 
& + \gamma \mathbb{E}_{(q,a)\sim\mathcal{D}_r}  \underbrace{\left[l_r\left(q \mid a; \left(\pi_{\theta} + \mathbf{B}_r\mathbf{A}\right)\right)\right]}_{\mathclap{\text{entropy-protected retention}}}  \\
& + \lambda \underbrace{\mathbb{E}_{x_t \in \mathcal{H}} \left[ \| \nabla_{\mathbf{A}} S_q(x_t) \|^2 \right]}_{\text{structural invariance}}.
\end{split}
\end{equation}

Inspired by the classic hinge loss, we reverse the optimization direction and define $\mathcal{L}_{\text{IHL}}$ which enables precise knowledge forgetting on low entropy tokens by suppressing target prediction probabilities while promoting next-best tokens. Retention loss $l_r$ strengthens core capabilities, and gradient constraints on $\mathbf{A}$'s high entropy tokens preserve structural integrity. Strict gradient isolation is enforced: each forgetting expert $\mathbf{B}_f^d$ updates via $\nabla l_f|_{\mathcal{L}d}$, $\mathbf{B}_r$ is updated via $\nabla l_{\text{IHL}}|_{\mathcal{L}r}$. Shared matrix $\mathbf{A}$ is updated solely via high entropy tokens ($\nabla S_q|_{\mathcal{H}}$).

\subsection{Inference Stage}

During inference, AsymUnlearn employs an entropy-aware conditional computation architecture to balance efficiency and accuracy. Given an input token $x_t$ in multi-forgetting tasks, real-time Tsallis entropy $S_q(x_t)$ computation triggers differentiated paths based on entropy thresholds:
\begin{equation}
\begin{split}
y = 
\begin{cases} 
\mathbf{W}_0x + \mathbf{A}x + \sum_{d=1}^k g_d(x_t)\mathbf{B}_f^d\mathbf{A}x & S_q(x_t)>1.2 \\
\mathbf{W}_0x + \mathbf{B}_{i^*}\mathbf{A}x & S_q(x_t)\leq1.2.
\end{cases}
\end{split}
\end{equation}

High entropy tokens use multi-expert fusion: aggregate $\mathbf{A}$ and top-3 $\mathbf{B}_f^d$ outputs via pre-trained gating weights $g_d(x_t)$. This design preserves structural integrity and cross-domain consistency. Low entropy tokens activate single-expert bypass: engage only the highest-weight $\mathbf{B}_{i^*}$, avoiding redundant computations and localization interference.

\begin{table*}[t]
\centering
\begin{tabular}{l|cccccccc|ccc}
\toprule
\textbf{Model/Category} & \multicolumn{8}{c}{\textbf{FT \& Model Editing}} &  \multicolumn{3}{c}{{\textbf{LoRA Variations$_{(r=8)}$}}} \\
\midrule
{\textbf{Llama3-8B}}  &  Base & RMU & ELM & GA & RL & NPO & NPO\_KL & NPO\_GD & LoRA & AsymLoRA & Ours  \\
\midrule
  WMDP-Bio↓ & 71.2 & 49.4 & 33.3 & \textbf{23.3} & 24.7 & 58.1 & 64.3 & 56.2 & 28.7 & 25.7 & \underline{24.4}  \\
 \textit{WMDP-Cyber}↓ & \textit{45.3} & \textit{37.0} & \textit{26.6} & \textit{\textbf{24.0}} & \textit{26.6} & \textit{34.4}  & \textit{41.3} & \textit{33.1} & \textit{32.1} & \textit{28.8} & \textit{\underline{25.6}}   \\
  MMLU↑ & 62.1 & 40.1 & \underline{57.2} & 24.8 & 23.0 & 50.1  & 56.0 & 51.9 & 39.6 & 55.3 & \textbf{57.8}  \\
  \textit{Flu-mean}↑ & \textit{2.97} & \textit{2.96} & \underline{\textit{3.07}} & \textit{1.00} & \textit{1.00} & \textit{3.07} & \textit{2.97} & \textit{3.03} & \textit{2.23} & \textit{2.23} & \textit{\textbf{3.46}} \\
  Flu-var↓ & 1.91 & 1.88 & 2.18 & 0.00 & 0.00 & 1.86  & 1.96 & 2.08 & 1.42 & 1.42 & \textbf{1.17}  \\
\midrule
{\textbf{Zephyr-7B}}  &  Base & RMU & ELM & GA & RL & NPO & NPO\_KL & NPO\_GD & LoRA & AsymLoRA & Ours  \\
\midrule
  WMDP-Bio↓ & 64.4 & 30.2 & 29.6 & 24.7 & 24.0 & 63.5 & 64.3 & 63.5 & 34.2 & 27.1 & \textbf{24.4} \\
  \textit{WMDP-Cyber}↓ & \textit{44.3} & \textit{27.3} & \textit{27.2} & \textit{26.8} & \textit{24.7} & \textit{43.6}  & \textit{45.3} & \textit{43.1} & \textit{32.1} & \textit{26.3} & \textbf{\textit{24.0}}  \\
  MMLU↑ & 58.5 & \underline{57.8} & 56.2 & 23.0 & 26.4 & 57.8  & 57.4 & \textbf{58.0} & 37.4 & 54.1 & 56.4  \\
  \textit{Flu-mean}↑ & \textit{2.97} & \textit{2.92} & \underline{\textit{2.99}} & \textit{1.00} & \textit{1.00} & \textit{2.98}  & \textit{2.95} & \textit{2.93} & \textit{2.47} & \textit{2.47} & \textbf{\textit{3.11}}  \\
  Flu-var↓ & 1.98 & 2.03 & 2.00 & 0.00 & 0.00 & 2.12  & 1.91 & 2.08 & 1.57 & 1.57 & \textbf{1.33}  \\

\bottomrule
\end{tabular}
\caption{\textbf{Multiple-choice accuracy of five LLMs on the WMDP benchmark (forget) and the full MMLU (retain) after unlearning.} Our method enables model-agnostic unlearning, reducing WMDP accuracy to around 25\% while preserving MMLU performance, and achieves stable forgetting with superior fluency by avoiding entangled errors in knowledge domains.}
\label{tab:WMDP_performance}
\end{table*}

\begin{figure*}[ht]
\centering
\includegraphics[scale=0.50]{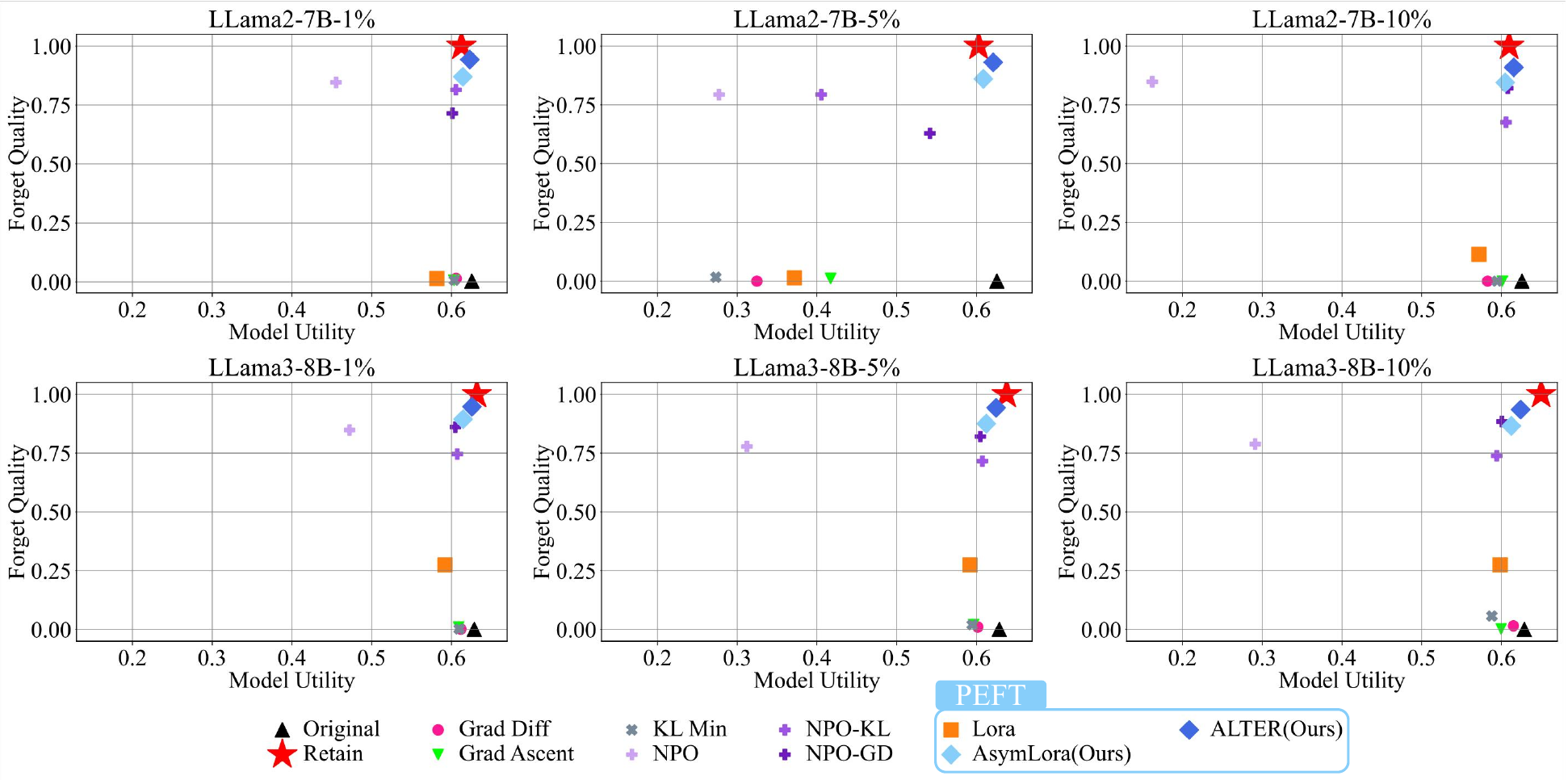}
\caption{Utility-forgetting trade-off at 1\%/5\%/10\% unlearning ratios for Llama2-7B (top) and Llama3-8B (bottom). GradDiff/Ascent and KLMin show low forgetting efficacy or severe utility loss. NPO incurs utility drops. Standard LoRA maintains utility but minimal forgetting gain. Our AsymLoRA/ALTER achieve near-complete forgetting with Retain-matched utility.}
\label{tofu_exp}
\end{figure*}

\begin{figure}[h]
\centering
\includegraphics[scale=0.45]{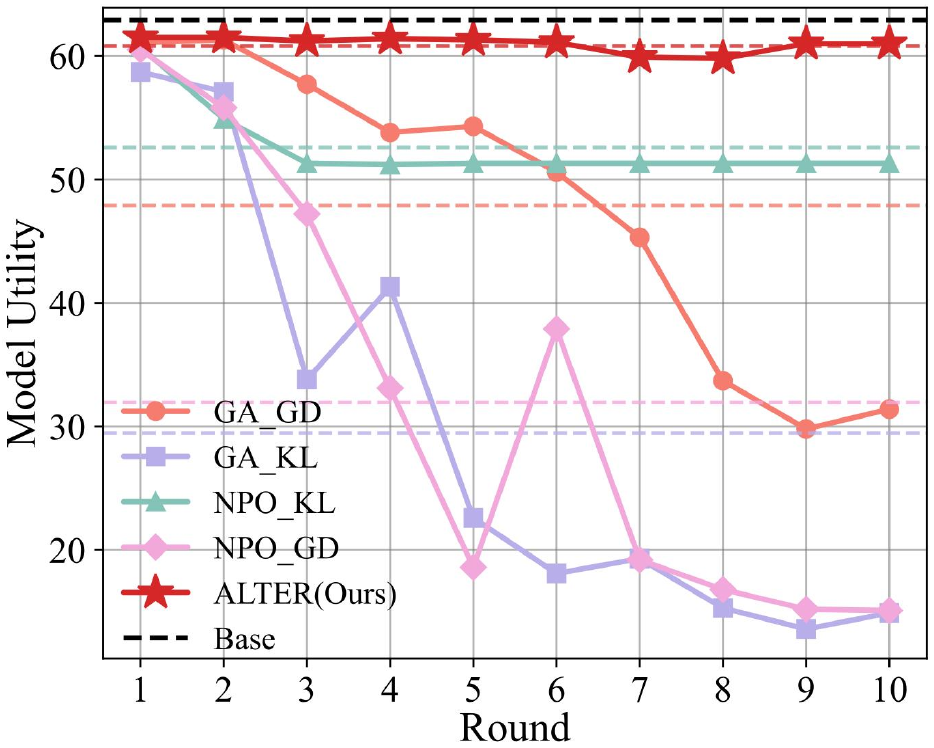}
\caption{Average model utility of baselines across Sequential Unlearning rounds for TOFU-injected Llama3-8B, with the forgetting set expanded from 1\% to 10\%.} 
\label{continueUnlearn}
\end{figure}

\section{Experiment}
\subsection{Experiment Setting}
\paragraph{Datasets}
% We evaluate using three benchmarks: TOFU are \cite{maini2024tofu} 200 synthetic author profiles (4k QA pairs; unlearning sets: 1\%,5\%,10\%), WMDP \cite{li2024wmdp} are sensitive knowledge: biosafety/cybersecurity/ chemical safety (no training set for chemistry, we did not use it.), and MUSE-HarryPotter \cite{rowling2023harry} (copyright-focused book unlearning). Additionally, to assess the general capabilities of LLMs, we include the fact-based question answering benchmark MMLU \cite{hendrycks2021measuring}.

We evaluate using three benchmarks: TOFU \cite{maini2024tofu}, which consists of 200 synthetic author profiles (4k QA pairs) with unlearning sets of 1\%, 5\%, and 10\%; WMDP \cite{li2024wmdp}, which assesses knowledge in sensitive domains such as biosafety, cybersecurity, and chemical safety (note: there is no training set for the chemistry subset, so we did not use it); and MUSE-HarryPotter \cite{rowling2023harry}, a copyright-focused benchmark for book unlearning. Additionally, to evaluate the general capabilities of LLMs, we include the fact-based question answering benchmark MMLU \cite{hendrycks2021measuring}.

\paragraph{Baselines and Backbones}
For the selection of baseline methods and backbone models, we strictly follow the default recommendations provided by each benchmark.

For TOFU, we use Llama3-8B and Llama2-7B \cite{touvron2023llama} as the base models. Baselines include Grad. Diff~\cite{liu2022continual}, Gradient Ascent (GA)~\cite{thudi2022unrolling}, KL Min~\cite{nguyen2020variational}, NPO, NPO\_KL, and NPO\_GD~\cite{zhang2024negative}.

For WMDP, we use Zephyr-7B~\cite{tunstall2023zephyr} and Llama3-8B~\cite{dubey2024llama} as the base models. The baselines are Representation Misdirection for Unlearning (RMU) \cite{maini2024tofu}, Erasure of Language Memory (ELM)~\cite{gandikota2025erasing}, GA, Random Label (RL)~\cite{yao2024machine}, NPO, NPO\_KL, and NPO\_GD. 

For HarryPotter, we use Llama-2-7B~\cite{touvron2023llama} as the base model and compare it with six baselines: WHP~\cite{liu2024revisiting}, ELM, GA, GD, KL, and NPO.

\paragraph{Evaluation Metrics}
We verify method effectiveness via multiple established metrics, and compute their harmonic mean for comprehensive statistical results.

For TOFU, forget quality uses a KS test p-value (higher means greater distribution similarity between unlearned/retained models). Model utility assesses retention set and real-world performance. Metric details in Appendix E.2.1.

For WMDP, forget quality is evaluated via multiple-choice accuracy on bio/cybersecurity questions. Model utility is assessed on MMLU through multiple-choice accuracy.

For HarryPotter, forget quality is evaluated via BLEU \cite{papineni2002bleu} and ROUGE-L \cite{lin2004rouge} between ground-truth and generated completions of 200-token prefixes. Model utility is evaluated on MMLU.

\begin{table}[t]
\centering
\begin{tabular}{l|ccc|cc}
\toprule
\multirow{3}{*}{\textbf{Method}} & \multicolumn{5}{c}{\textbf{Harry Potter}} \\
 & \multicolumn{3}{c}{Forget Perf.} & \multicolumn{2}{c}{Retain Perf.} \\
 & {BLEU} & {R-L} & ASG↓ & {MMLU↑} & {Ful.↑} \\
\midrule
Original & 74.8 & 85.1 & 74.1 & 46.3 & 4.0 \\
Retain   & 1.9  & 9.8 &  0  & 47.8 & 2.8  \\
\midrule
Fine-tune & 6.4 & 17.2 & 5.9 & 46.0 & 1.9 \\
GA & 0 & 0 & 6.0 & 26.9 & 1.0 \\
GD & 3.9 & 14.5 & 3.4 & 43.6 & 1.8 \\
NPO & 1.5 & 5.3 & 2.5 & 42.7 & 2.9 \\
KL & 1.2 & 8.9 & \textbf{0.8} & 41.1 & \underline{3.1} \\
WHP & 23.6 & 17.9 & 14.9 & \underline{44.4} & 2.5 \\
ELM & 8.1 & 9.0 & 2.7 & \textbf{44.6} & 2.8 \\
\midrule 
LoRA & 7.2 & 11.5 & 3.5 & 38.9 & 2.3 \\
A-LoRA & 5.9 & 10.4 & 1.9 & 43.8 & 2.3 \\
Ours & 4.7 & 9.6 & \underline{1.3} & \textbf{44.6} &  \textbf{3.3} \\
\bottomrule
\end{tabular}
\caption{Performance on HarryPotter dataset. R-L and Ful. denote the ROUGE-L score and fluency-mean, respectively. A-LoRA denotes the AsymLoRA. Instead, average similarity gap (ASG) serves as the target for forget performance.}
\label{tab:HP_performance}
\end{table}

We evaluate generated content fluency via GPT-4o (average of 5 responses). While not perfectly aligned with human judgment, this approach provides a practical solution \cite{li2024single,shi2024detecting}. For WMDP, fluency assessment uses open-ended reasoning generation instead of direct ABCD option scoring. Details in Appendix E.2.4.

\paragraph{Configurations}
The experiment configurations are as follows: learning rate is $\eta_B$ = $10^{-3}$ and $\eta_A$ = $10^{-5}$, the weights for the forget loss and retain loss are set to $\beta$ = 1.0, $\gamma$ = 1.0, $\lambda$ = 0.01, batch size = 4, epoch = 3. 
The hardware and software configurations used in our experiments are as follows.
CPU: Intel(R) Xeon(R) Platinum 8468V, 800MHZ, 48cores;
GPU: NVIDIA TESLA H800 80 GB;
Operating system: Ubuntu 20.04;
Deep learning framework: PyTorch 2.4.1.

\subsection{Main Results}
\paragraph{Hazardous Knowledge Unlearning}
% \noindent \textbf{Hazardous Knowledge Unlearning}\ \ \
Our method achieves strong model-agnostic unlearning: on WMDP (Tab.\ref{tab:WMDP_performance}), it reduces biosafety/cybersecurity accuracy to near-random ($\sim$25\%) while preserving full MMLU performance across models. While GA/RL methods achieve comparable WMDP reduction, they catastrophically degrade MMLU performance to 23-26\%. AsymLoRA and our approach employ asymmetric architectures for dual parameter isolation—between unlearning subtasks and between unlearning and retention tasks—enabling effective forgetting without utility compromise.
Furthermore, our method achieves stable high-precision forgetting. On Llama3-8B/Zephyr-7B, it shows the highest Flu-mean (3.46/3.11) and lowest Flu-var (1.17/1.33), indicating superior generation consistency. In contrast, RMU/ELM reduce WMDP scores but suffer lower fluency, while NPO variants degrade MMLU performance. This suggests existing methods introduce entangled errors in related knowledge domains. We avoid this by preserving shared knowledge via high entropy tokens, ensuring precise removal while maintaining knowledge integrity.

\paragraph{Entity Unlearning}
The TOFU entity unlearning results (Fig.\ref{tofu_exp}) demonstrate that AsymLoRA/ALTER achieves near-perfect forget quality while maintaining Retain-utility across architectures and forgetting ratios, a performance superiority attributable to our architectural innovation. Unlike gradient-based methods that induce severe capability degradation through destructive parameter updates, or NPO variants that exhibit compromised forget quality due to blunt regularization, our framework overcomes standard LoRA's rigidity via entropy-driven token-level isolation. This enables surgical knowledge removal while preserving structural integrity, evidenced by the approximation result with the Retain model's performance. Crucially, these results validate our core solution: asymmetric knowledge partitioning fundamentally decouples forgetting precision from capability preservation, achieving Pareto-optimal balance through targeted parameter isolation at minimal computational cost.

\paragraph{Copyrighted Unlearning}
The HarryPotter copyrighted unlearning results are shown in Tab.\ref{tab:HP_performance}. We achieve text similarity scores comparable to the Retain model. The high-low entropy mechanism effectively distinguishes copyrighted entities from functional words, maintaining the Model Utility close to original levels. Strong baselines KL minimization and ELM demonstrate competitive ASG and general utility performance but exhibit noticeable fluency reduction after decoupled training. The GD method alleviates model collapse yet still incurs significant MMLU degradation. In contrast, our approach achieves near-optimal ASG while preserving superior MMLU and fluency performance, demonstrating the balanced capability of the entropy-based strategy.

\subsection{Analysis}
\paragraph{Continue Unlearning}
As shown in Fig.\ref{continueUnlearn}, ALTER demonstrates exceptional stability in model utility preservation during continuous unlearning, maintaining performance near base model levels with minimal degradation. In contrast, baseline methods exhibit progressive deterioration: gradient-based approaches (GA\_GD, GA\_KL) show severe utility loss, while NPO variants display moderate degradation. Among baselines, NPO\_KL preserves utility most effectively yet remains substantially inferior to ALTER. 

\begin{figure}[t]
\centering
\includegraphics[scale=0.45]{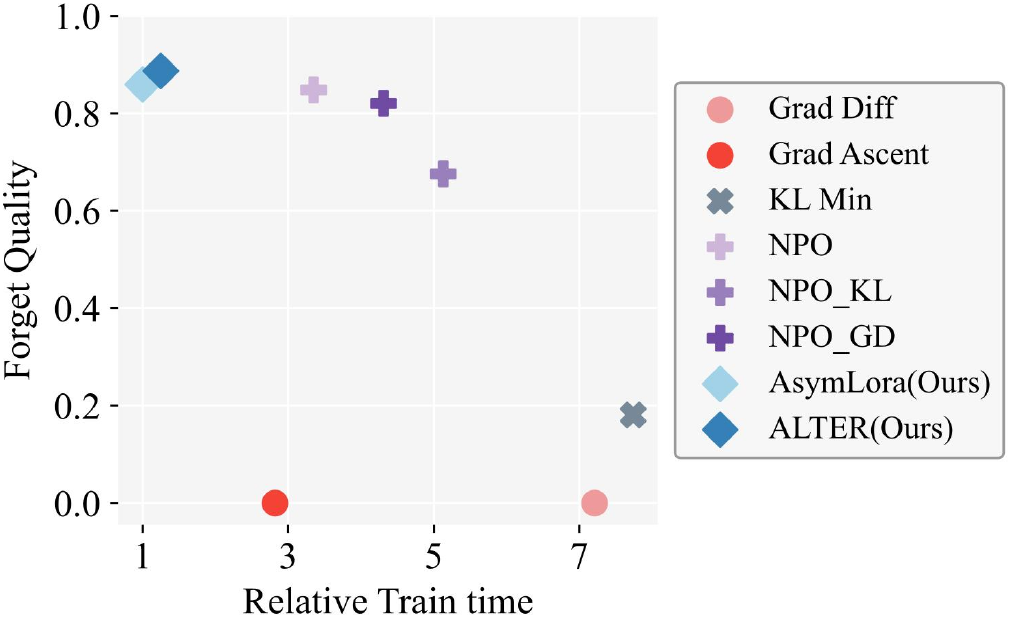}
\caption{Trade-off between forget quality and relative training time for Llama2-7B on TOFU-10\%. The top-left corner indicates better forgetting performance and efficiency.} 
\label{RE}
\end{figure}

\paragraph{Unlearning Time Efficiency}
Time efficiency is critical for LLM unlearning, especially compared to retraining from scratch.
As shown in Fig.\ref{RE}, our method reduces time costs by 86.1\% to 87.1\% using the AsymLoRA framework, achieving orders-of-magnitude higher forgetting quality. Baseline LoRA harms model utility, while AsymLoRA achieves near-optimal quality with minimal time cost (unit: 1.0). ALTER further improves performance at a modest 1.25× cost. By decoupling shared learning of high entropy tokens and targeted forgetting of low entropy tokens, our approach minimizes computation while maximizing unlearning efficacy, offering efficient solutions for large-scale models.

\subsection{Ablation Study}
All LoRA variants use rank$=$8 to balance effectiveness and performance. Details of different ranks for LoRA and the positions of decomposition modules are provided in Appendix G.

\section{Related Work}

\paragraph{Unlearning}
% \noindent \textbf{Unlearning}\ \ \ 
% Machine unlearning has emerged as a promising approach in LLM alignment research \cite{liu2025rethinking, thaker2025position}, aiming to address privacy, security, and bias issues in large language models \cite{xu2025obliviate}. Within unlearning methods, gradient ascent is a ``parameter-optimization'' technique that counteracts the influence of target knowledge by reversing the optimization direction of model parameters.
GD \cite{liu2022continual, thaker2025position} applies gradient ascent to make the model ``forget'' specific content and introduces an additional loss term to constrain the deviation between the unlearned and original models. \cite{yu2023unlearning} identifies bias-related neurons using Integrated Gradients and performs gradient ascent only on these neurons\cite{liu2025rethinking}, minimizing impact on other model capabilities. \cite{wangetal2023kga} balances forgetting effectiveness with performance retention by minimizing the KL divergence between the ``unlearned model'' and the ``original model'' on non-forgetting corpora. The SOUL algorithm \cite{jiaetal2024soul} optimizes the parameter update direction using second-order information approximated from the Hessian matrix, achieving unlearning while preserving model utility.
Recently, such as RMU \cite{dang2025effects} and LUNAR \cite{shen2025lunar}, use methods similar to guided vectors \cite{cao2024personalized, cha2024towards} and dedicated UNL tokens \cite{yu2025unierase} to locate and modify local model parameters associated with target knowledge, thereby redirecting the model into an inability space. 
% Recently, \cite{cha2024towards} achieved efficient unlearning by customizing the LoRA loss function during fine-tuning, without compromising forgetting quality.

\paragraph{Multi-LoRA Architecture}
% \noindent \textbf{Multi-LoRA Architecture}\ \ \ 
LoRA \cite{hu2022lora} leverages low intrinsic dimensionality \cite{aghajanyan2020intrinsic} via trainable low-rank matrices in frozen models, efficiently approximating gradient updates. Its low latency and performance drove wide adoption. Subsequent multi-LoRA variants enhance efficiency and stability: \cite{huang2024lorahub} uses domain-adaptive adapter combinations; \cite{wang2023multilora} reduces parameter dependency via horizontal scaling. These collectively advance hybrid LoRA architectures \cite{zadouri2024pushing}.
MoE-LoRA \cite{dou2023loramoe} integrates LoRA with Mixture-of-Experts (MoE) to reduce multi-task interference via task-specific adapters. HydraLoRA \cite{tian2024hydralora} employs asymmetric LoRA with automatic clustering and MoE for efficient adaptation.
% on heterogeneous corpora.

\paragraph{Entropy}
Entropy has been widely applied across various areas of LLMs: semantic entropy measures information density via uncertainty assessment \cite{guo-etal-2025-hash}; entropy regularization penalizes low entropy predictions \cite{miller2002global, pereyra2017regularizing} while maximizing prediction entropy \cite{setlur2022maximizing}, enhancing adversarial robustness \cite{jagatap2022adversarially} and domain generalization \cite{zhao2020domain}. In unlearning, methods constrain entropy by maximizing \cite{peer2022improving, jha2025entropy} on forgetting sets to reduce target confidence and minimizing on retention sets to preserve discriminative ability \cite{entesari2025constrained, tarun2023fast, jung2025entun}.

\section{Conclusion}
In this work, we propose ALTER, a novel and universal method for LLM unlearning, establishing a new paradigm for parameter-efficient unlearning. ALTER leverages an asymmetric LoRA structure and token entropy to establish a dynamic forgetting boundary that preserves the integrity of the model's knowledge topology while eliminating target information.
By decoupling the unlearning process from the LLMs' billions of parameters, ALTER delivers an efficient unlearning framework. With minimal side effects, it maintains model utility comparable to the Retain model across varying architectures. Our experiments across three unlearning tasks validate ALTER’s effectiveness, setting a foundation for responsible AI deployment in real-world scenarios. 

\section{Acknowledgments}
This research is supported by the National Natural Science Foundation of China (Grants 62306067, 62402093, W2433163), Sichuan International Science and Technology Innovation Cooperation Project (ID 2024YFHZ0317), Sichuan Science and Technology Program (ID 2025ZNSFSC0479), Chengdu Science and Technology Bureau Project (ID 2024-YF09-00041-SN), the Postdoctoral Fellowship Program (Grade C) of the China Postdoctoral Science Foundation (Grant GZC20251053), and Huawei Funding (ID H04W241592; partially supported by UESTC Kunpeng\&Ascend Center of Cultivation).

\bibliography{aaai2026}
\end{document}